%% file: main_arxiv_simplefigs.tex
\let\mypdfximage\pdfximage
\def\pdfximage{\immediate\mypdfximage}
\documentclass{article}

\usepackage[nonatbib,preprint]{neurips_2020}

\usepackage{titling}
\usepackage{graphicx}
\usepackage{booktabs} 
\usepackage{amsfonts}
\usepackage{amsmath}
\usepackage{subcaption}
\usepackage{tcolorbox}
\usepackage{adjustbox}
\usepackage{placeins}
\usepackage{capt-of}
\usepackage{mdframed}
\usepackage[percent]{overpic}
\usepackage{hyperref}
\usepackage{wrapfig}

\newcommand{\Section}[1]{\vspace{-0pt}\section{#1}\vspace{-0pt}}
\newcommand{\Subsection}[1]{\vspace{-0pt}\subsection{#1}\vspace{-0pt}}

\DeclareMathOperator*{\argmin}{arg\,min}

\graphicspath{{./arxivfigs/}}
\title{Deep S$^3$PR: Simultaneous Source Separation and Phase Retrieval Using Deep Generative Models (Long Version)}

\author{%
	Christopher A.~Metzler \& Gordon Wetzstein \\
	Stanford University \\
	\texttt{cmetzler@stanford.edu} \& \texttt{gordon.wetzstein@stanford.edu}\\
}
\date{}

\begin{document}

\maketitle

\begin{abstract}
This paper introduces and solves the simultaneous source separation and phase retrieval (S$^3$PR) problem. S$^3$PR is an important but largely unsolved problem in a number application domains, including microscopy, wireless communication, and imaging through scattering media, where one has multiple independent coherent sources whose phase is difficult to measure. 
In general, S$^3$PR is highly under-determined, non-convex, and difficult to solve. In this work, we demonstrate that by restricting the solutions to lie in the range of a deep generative model, we can constrain the search space sufficiently to solve S$^3$PR. 

Code associated with this work is available at \url{https://github.com/computational-imaging/DeepS3PR}.
\end{abstract}

\Section{Introduction}
\label{sec:Intro}
\input{sections/Intro.tex}

\Section{Related work}
\label{sec:RelatedWork}
\input{sections/RelatedWork.tex}

\Section{Naive S$^3$PR solutions}
\label{sec:AltMethods}
\input{sections/SSSPRwClassical_simplefigs.tex}
\input{sections/SSSPRwDiscriminative.tex}

\Section{S$^3$PR using deep generative models}
\label{sec:SSSPRwGen}
\input{sections/SSSPRwGenModel.tex}

\Section{Experimental results}
\label{sec:ExpResults}
\input{sections/ExpResults_simplefigs.tex}

\Section{Discussion}
\label{sec:Discussion}

\input{sections/Discussion.tex}

\begin{ack}
C.M.~was supported by an appointment to the Intelligence Community Postdoctoral Research Fellowship Program at Stanford University administered by Oak Ridge Institute for Science and Education (ORISE) through an interagency agreement between the U.S. Department of Energy and the Office of the Director of National Intelligence (ODN). G.W.~was supported by an NSF CAREER Award (IIS 1553333), a Sloan Fellowship, and a PECASE by the ARL.
\end{ack}

\bibliographystyle{./myIEEEtran}
\bibliography{myref}

\newpage
\title{Supplement to Deep S$^3$PR: Simultaneous Source Separation and Phase Retrieval Using Deep Generative Model}
\maketitle
\input{sections/Supp_Sec_simplefigs.tex}

\end{document}

%% file: sections/Intro.tex
We define the S$^3$PR problem as follows. Recover the signals $x_1,~x_2,~\ldots,~x_L\in\mathcal{C}$, where $\mathcal{C}$ is some subset of $\mathbb{R}^n$ (e.g.~the set of natural images), from a noisy measurement vector $y\in\mathbb{R}^m$ given by
\begin{align}\label{eqn:SSSPR}
y = \sum_{l=1}^L |\mathbf{A}_lx_l|^2+w,
\end{align}
where the square is elementwise, the measurement matrices $\mathbf{A}_l\in\mathbb{C}^{m\times n}$ are known for all $l$, and $w$ represents additive noise. 
In the remainder, we focus on the special case $\mathbf{A}_1=\mathbf{A}_2=\ldots=\mathbf{A}_L$.

For $L=1$ this problem is just standard phase retrieval (PR), a problem for which myriad solutions exist~\cite{gerchberg1972practical,fienup1978reconstruction, HIO, griffin1984signal}. However, when $L>1$ things become significantly more challenging. In particular, one is forced to disentangle the components of $y$ that came from $x_i$ from the components that came from $x_j$, with $i\neq j$. That is one must solve a source separation (SS) problem as well as multiple PR problems.

\paragraph{Motivation}
S$^3$PR is intrinsic to a variety of different application domains where one measures the intensity of a field formed by multiple independent coherent sources. 
Under these conditions, the {\em fields} within a single source add but the {\em intensities} between distinct sources add. 
This situation appears in partially coherent phase imaging microscopy~\cite{LauraRelatedWork1,LauraRelatedWork2}, correlation-based imaging through thin scattering media and around corners with highly separated or multi-spectral objects~\cite{bertolotti2012non,katz2014non,li2019single,viswanath2018indirect,metzlerOptica}, transmission-matrix based imaging through thick scattering media with multiple independent sources~\cite{rajaei2016intensity,sharma2019inverse}, and even multiple source localization with mmWave 5G~\cite{hassanieh2018fast}. A detailed description of S$^3$PR's role in correlation-based imaging through scattering media is provided in the supplement.

Despite the prevalence of the S$^3$PR problem, no general solution to to S$^3$PR yet exist. 
The lack of existing solutions is likely because S$^3$PR is simply too non-convex and under-determined to be solved with conventional algorithms. While a cascaded solution, that is SS followed by PR, could work in  principle, Section~\ref{sec:ExpResults} demonstrates that in practice the components $|\mathbf{A}x_1|^2,~|\mathbf{A}x_2|^2,~\ldots,~|\mathbf{A}x_L|^2$ are too similar to reliably separate with traditional SS algorithms.

\paragraph{Our contribution}
In this work we solve the S$^3$PR problem by imposing strong but realistic priors on the reconstructed signals. In particular, we constrain $x_1,~x_2,~\ldots,~x_L$ to lie in the range of a generative neural network $G : \mathbb{R}^k\rightarrow \mathbb{R}^n$ with $k\ll n$. That is, $\mathcal{C}=\text{Range}(G)$ and for all $x_l\in\mathcal{C}$ there exists some latent vector $z_l\in\mathbb{R}^k$ such that $x_l = G (z_l)$.

Using this constraint, we recover $x_1,~x_2,~\ldots,~x_L$, with our estimates $\hat{x}_l=G(\hat{z}_l)$, from $y$ by solving the optimization problem
\begin{align}
\hat{z}_1,\hat{z}_2,\ldots,\hat{z}_L = \argmin_{z_1,z_2,\ldots,z_L}{\big \|}y - \sum_{l=1}^L|\mathbf{A}G(z_l)|^2{\big \|}^2_2,
\end{align}
using an alternating descent algorithm.

To our knowledge, this work represents the first general solution to S$^3$PR. Accordingly, it opens up a variety of application domains. In Section \ref{sec:ExpResults} we apply this method to simulated Gaussian, coded-diffraction-pattern (CDP), and Fourier measurements and demonstrate the successful recovery of numbers ($\mathcal{C}=$~MNIST dataset) and articles of clothing ($\mathcal{C}=$~Fashion MNIST dataset~\cite{xiao2017fashion}).

\paragraph{Limitations} 
Our results have a few limitations. 
First, our present reconstructions are low resolution and come from fairly restrictive classes --- digits and articles of clothing. Second, while we provide extensive evidence that deep generative models can be used to solve S$^3$PR, our current results are purely empirical. S$^3$PR remains a highly under-determined and non-convex problem and deriving the conditions under which it can and cannot be solved remains an important but open problem.

%% file: sections/RelatedWork.tex
While the S$^3$PR problem is new, both PR and SS have been studied extensively and have a vast literature. 
Likewise, while not previously used for S$^3$PR, deep generative models have recently been applied to a range of inverse problems. 
We now highlight a few of the most prominent of these works.

\Subsection{Phase retrieval}
The optics community has studied the PR problem continuously since the 1970s~\cite{gerchberg1972practical,fienup1978reconstruction, HIO, griffin1984signal,ERHIO, OSS}. 
In the last decade, PR has caught the attention of the optimization and machine learning community as well~\cite{candes2013phaselift,candes2015phase,goldstein2016phasemax,bahmani2017phase}. For a benchmark study of over a dozen popular PR algorithms, see PhasePack \cite{chandra2017phasepack}. No existing PR algorithm can handle S$^3$PR --- they are designed for a fundamentally different forward model that does not mix measurements.

\paragraph{Compressive phase retrieval}
\cite{moravec2007compressive} recognized that if one imposed a prior on the reconstruction one could perform PR using significantly fewer measurements. This initial work has been followed up by numerous others that have imposed more and more elaborate priors on the reconstruction~\cite{mukherjee2012iterative,prGAMP,dolphin,bm3dprgamp}.  Again, none of these algorithms can handle S$^3$PR.

\paragraph{Multi-source phase retrieval}
A handful of works~\cite{chern2002blind,guo2018multi,guo2019multiple,guo2019blind} have studied multi-source PR, defined as recovering $x_1,~x_2,~...,~x_L$ from 
\[
y = |\sum_{l=1}^L \mathbf{A}_lx_l|^2.
\]
This model resembles Equation \eqref{eqn:SSSPR}, but note that here the mixing occurs {\em before} the non-linearity. 
This difference makes the reconstruction problem significantly easier as it enables the application of a standard PR algorithm to recover $\sum_{l=1}^L \mathbf{A}_l x_l$ followed by a standard SS algorithm to recover $x_1,~...,~x_L$. In contrast, the S$^3$PR problem defined by Equation \eqref{eqn:SSSPR} does not lend itself to similar cascaded solutions as existing SS algorithms struggle to separate $|\mathbf{A}x_1|^2$ and $|\mathbf{A}x_2|^2$ when $x_1$ and $x_2$ are drawn from the same distribution/class; see the reconstructions in Section~\ref{sec:ExpResults}.

\paragraph{Phase retrieval and blind demodulation}
In various imaging applications, one records measurements of a signal $x_1$ that is illuminated by an unknown signal $x_2$. In this context, the measurement model becomes 
\[
y = | \mathbf{A} (x_1 \circ x_2)|^2,
\]
where $\circ$ denotes the Hadamard (elementwise) product. While solutions to this problem exist~\cite{kane2008principal,bendory2019blind}, the fact that mixing occurs before the non-linearity makes the problem fundamentally different from S$^3$PR.

\paragraph{Partially coherent phase retrieval and source recovery}
When one images a fixed object $x$ with multiple unknown Fourier plane (K\"{o}hler) illumination sources, one is presented with a constrained S$^3$PR problem
\[
y = \sum_{l=1}^L| \mathbf{A} (x_l)|^2\text{ with }x_l=x\circ \mathbf{F} e_l,
\]
where $e_1$, $e_2$, ... $e_L$ are unknown 1-hot vectors representing the location of an illumination source and $\mathbf{F}$ is a 2-D discrete Fourier transform matrix which models their propagation to the object plane. By capturing multiple measurements of this form at different stand-offs (which corresponds to selecting a particular measurement matrix $\mathbf{A}$) one can reframe this problem as deconvolution and solve it using conventional algorithms~\cite{LauraRelatedWork1,LauraRelatedWork2}.

\Subsection{Source separation}

Traditional SS algorithms assume that the sources to be separated are statistically independent and that there are as many or more observations as there are unknowns~\cite{molgedey1994separation,comon1994independent,comon2010handbook}. When this is not the case, the problem is known as under-determined SS and is much more challenging.

Under-determined SS, which features prominently in reflection removal and hyperspectral imaging, has been accomplished by imposing priors on the reconstructed signals, for instance that they are sparse~\cite{bofill2001underdetermined,li2004analysis,takigawa2004performance,li2006underdetermined,levin2007user} or group sparse~\cite{drumetz2019hyperspectral} in some basis 
or have other structural properties that can be exploited~\cite{deleforge2017phase}. 
Under-determined SS can also be accomplished with convolutional neural networks~\cite{fan2017generic,zhang2018single,luo2019conv}.

\Subsection{Inverse problems with deep generative models}
Deep generative models have been used to solve a variety of under-determined inverse problems.

\paragraph{Compressive sensing}
The idea of solving the compressive sensing problem by recovering the latent vectors of a deep generative model was first proposed in \cite{bora2017compressed}.  
This work has been followed up by a number of papers which have sought to improve the speed~\cite{manoel2017multi,shah2018solving,DeepCS,MLVAMP,latorre2019fast} and generalizability~\cite{dhar2018modeling,hussein2019image,asim2019invertible} of the method.

\paragraph{Phase retrieval}
PR with generative models was introduced in \cite{hand2018phase} and \cite{shamshad2018robust}. 
This method has since been accelerated in~\cite{hyder2019alternating} and applied to various PR problems in~\cite{shamshad2019deep,shamshad2019adaptive}.

We note in passing that optimizing the latent variables of untrained networks, following ideas proposed in \cite{DIP}, can also be used to impose priors in order to help solve the PR problem~\cite{jagatap2019phase,bostan2020deep}.

\paragraph{Other inverse problems}
For the sake of completeness, we note that optimizing the latent variables of a deep generative models has also been used to perform blind demodulation~\cite{hand2019global}, blind deconvolution~\cite{asim2018blind}, matrix decomposition~\cite{aubin2019spiked}, and source separation~\cite{kong2019single}. We also note that optimizing the latent variables of untrained neural networks has been used for a variety of image decomposition tasks, including source separation~\cite{Double_DIP}.

%% file: sections/SSSPRwClassical_simplefigs.tex
Here we describe two naive solution with which one might approach S$^3$PR. As we will demonstrate in Section \ref{sec:ExpResults}, the first approach is largely ineffective. The second does not generalize.
\Subsection{SS followed by PR}\label{ssec:Sequential}

Given measurements
\begin{align}
y = \sum_{l=1}^L |\mathbf{A}x_l|^2+w,\nonumber
\end{align}
one could first use an under-determined SS algorithm to estimate $|\mathbf{A}x_1|^2,~|\mathbf{A}x_2|^2,~\ldots,~|\mathbf{A}x_L|^2$ and then use a PR algorithm ($L$ times) to recover $x_1,~x_2,~\ldots,~x_L$ from these estimates. That is, one could solve potentially solve S$^3$PR by performing under-determined SS followed by PR.

\begin{wrapfigure}{R}{.48\linewidth}
	\centering
	\includegraphics[width=.48\textwidth]{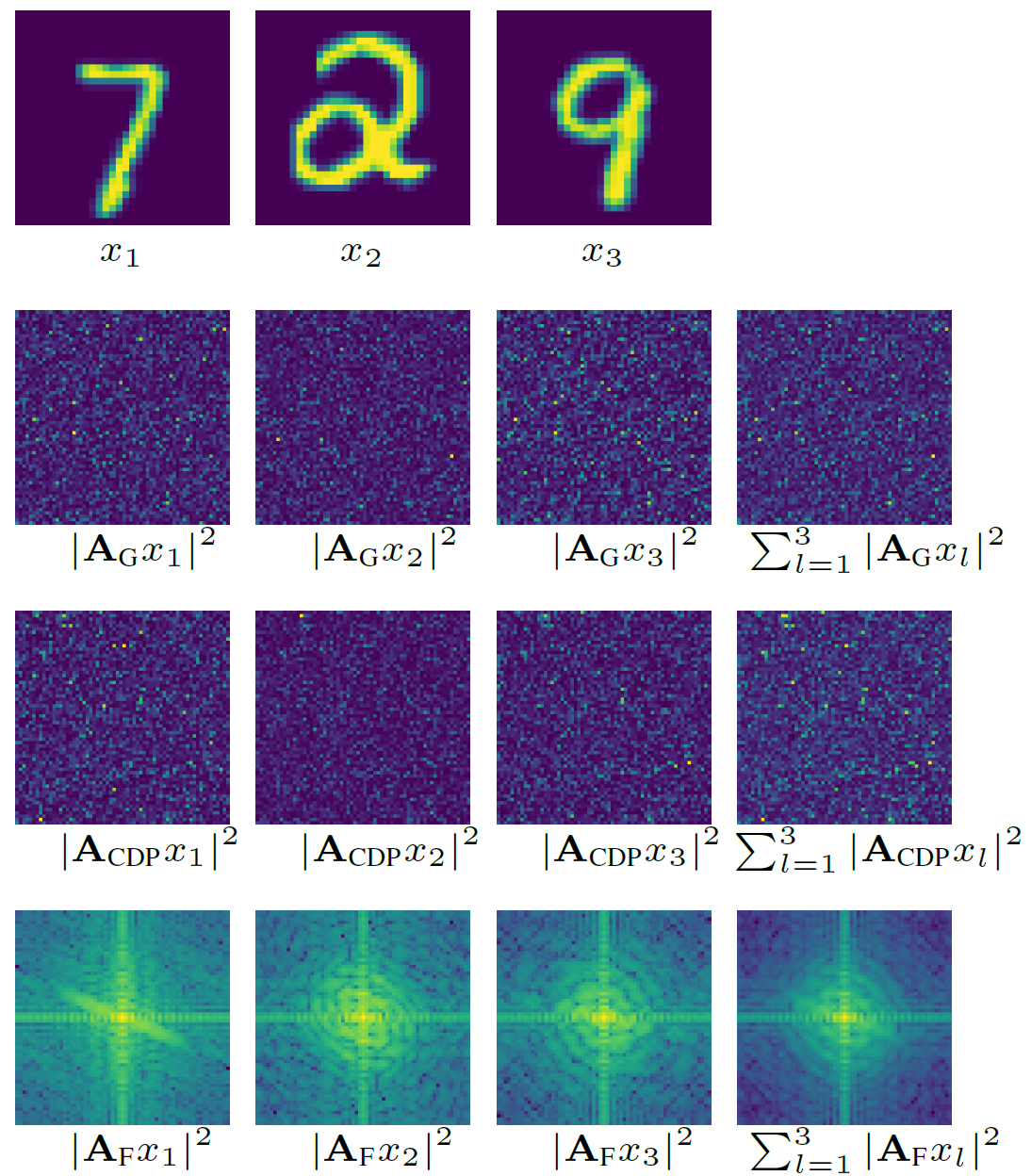}
	\caption{{\bf Visualization of measurements.} Examples of signals and their measurements with Gaussian ($\mathbf{A}_\text{G})$, CDP ($\mathbf{A}_\text{CDP}$), and Fourier ($\mathbf{A}_\text{F}$) measurement matrices. The Fourier measurements are displayed in log scale and the Gaussian and CDP measurements were reshaped from $64^2\times 1$ vectors into $64\times64$ images for visualization. The various measurements have little obvious structure and separating each of them from their sum is a challenging problem.}
	\label{fig:Measurements}
\end{wrapfigure}

The aforementioned SS step requires one to separate measurements   $|\mathbf{A}x_1|^2,~|\mathbf{A}x_2|^2,~\ldots,~|\mathbf{A}x_L|^2$ from their sum. Representative measurements and their sum are presented in Figure~\ref{fig:Measurements}. Separating such signals is a highly under-determined SS problem for which standard SS algorithms, like non-negative matrix factorization, do not apply. Instead, one needs to use prior information, e.g.~sparsity in some basis/dictionary, about the measurements to unmix them.

As a best attempt at dictionary-based under-determined SS, for each dataset and measurement matrix pair we learn a unique 500 element dictionary, $\mathbf{D}$ (for a total of 6 dictionaries). 
Each dictionary was formed using the K-SVD algorithm~\cite{KSVD} and was optimized to form one-hot representations of $60\,000$ training measurements, which were specific to the dataset and the measurement matrix. 

With a dictionary in hand, we then perform SS by finding an $L$-hot representation of $y$ in this dictionary using orthogonal matching pursuit (OMP)~\cite{OMP}. That is, we use OMP to approximately solve
\begin{align}\label{eqn:SS_subproblem}
\|y-\mathbf{D}\alpha\|_2^2 \text{ subject to } \|\alpha\|_0=L,
\end{align}
where $\|\alpha\|_0$ is the cardinality of $\alpha$. The result allows us to form the estimates
\begin{align}
\widehat{|\mathbf{A}x_l|^2}=D_l\alpha_l,
\end{align}
where $\alpha_l$ is the $l^{th}$ non-zero element of $\alpha$ and $D_l$ is the column of $\mathbf{D}$ (i.e., the element of the dictionary) associated with $\alpha_l$.

Following SS, we reconstruct $x_1,~x_2,~\ldots,~x_L$ by solving $L$ phase retrieval problems by applying gradient descent to the loss
\begin{align}\label{eqn:PR_subproblem}
\hat{x}_l = \argmin_x \|\widehat{|\mathbf{A}x_l|^2}-|\mathbf{A}x|^2\|^2_2.
\end{align}
We also experimented with the widely used Gerchberg--Saxton algorithm~\cite{gerchberg1972practical} but found that it ran far slower and offered worse performance. 

As we will demonstrate in Section \ref{sec:ExpResults}, this approach is largely ineffective at solving S$^3$PR.

%% file: sections/SSSPRwDiscriminative.tex
\Subsection{Discriminative neural networks}

One could potentially use discriminative deep neural networks to perform the source separation step within the above algorithm. One could even use a discriminative neural network to directly learn a mapping from mixed measurements, $y$, to images, $x_1,~x_2,\ldots,x_L$. However, with either approach such a network would become specific to the forward model, the number of signals in the mixture, the distribution of the signals in the mixture, and potentially the signal-to-noise ratio (SNR) -- obtaining the results in our paper would have required training 36 separate networks.

In contrast, our proposed method generalizes across forward models, SNRs, and the number of signals in the mixture -- we trained only two networks.

%% file: sections/SSSPRwGenModel.tex
A better way to perform S$^3$PR is to leverage deep generative models as priors.

\paragraph{Recovering latent variables}
We recover images $x_1,~x_2,~\ldots,~x_L$, with our estimates $\hat{x}_l=G(\hat{z}_l)$, from $y$ by solving the optimization problem
\begin{align}\label{eqn:Loss}
\hat{z}_1,\hat{z}_2,\ldots, \hat{z}_L = \argmin_{z_1,z_2,\ldots,z_L}{\big \|}y - \sum_{l=1}^L|\mathbf{A}G(z_l)|^2{\big \|}^2_2,
\end{align}
using an alternating descent algorithm. 
That is, we iteratively compute the loss~\eqref{eqn:Loss} and take a gradient step (with momentum) with respect to $z_1$, then compute the loss and take a gradient step (with momentum) with respect to $z_2$, etc. 
In practice, we found alternating descent (using the ADAM optimizer~\cite{kingma2014adam}) provided a near monotonic reduction of the loss and ran in two minutes on an Nvidia Titan RTX GPU.

For time/resource-sensitive applications, one could also use an alternating projection algorithm~\cite{shah2018solving,hyder2019alternating} or more advanced methods like AMP~\cite{manoel2017multi,MLVAMP} or ADMM~\cite{latorre2019fast}, which can provide significantly faster convergence.

\paragraph{Generative model}
\begin{figure}
	\centering
	\includegraphics[width=.6\columnwidth]{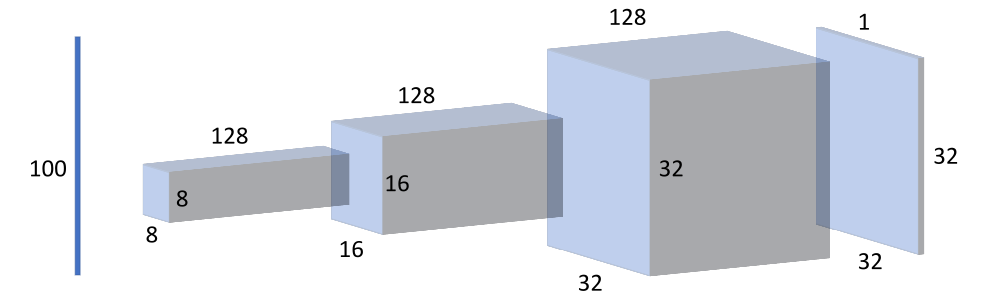}
	\caption{{\bf Simplified DC-GAN network architecture.} A simple convolutional neural network, which maps a 100 dimensional latent vector to a $32\times32$ image, serves as our generative model.}
	\label{fig:DCGAN}
	\vspace{-15 pt}
\end{figure}

Following~\cite{bora2017compressed} we use a variation of the DC-GAN~\cite{DCGAN} architecture as our deep generative model, $G(z)$. Our generator network is illustrated in Figure \ref{fig:DCGAN}. It takes a 100 dimensional latent vector, $z\in\mathbb{R}^{100}$, as an input and applies a fully connected layer followed by a reshaping operation and batch normalization to form an $8 \times 8\times 128$ dimensional feature map. This feature map is then upsampled $2\times$, convolved with $128$ different $3\times3$ filters, batch normalized, and feed through a leaky ReLU to form a $16\times16\times 128$ feature map. This feature map is then upsampled, convolved with $64$ different $3\times3$ filters, batch normalized, and feed through another leaky ReLU to form a $32\times32\times64$ feature map. Finally, this feature map is convolved with a single $3\times3$ filter and passed through a tanh nonlinearity to form the generated image.  

We train one such DC-GAN network to produce MNIST digits and another to produce Fashion MNIST articles of clothing. Each network was based off the Pytorch implementation of DC-GAN from \cite{PyTorch-GAN} and was trained using the code's default parameters. The networks were trained using the training portion of their respective datasets and tested, as described in the next section, on a subset of the testing portion.

%% file: sections/ExpResults_simplefigs.tex
We now apply the two proposed methods to Gaussian, coded diffraction pattern (CDP), and Fourier measurements of images of numbers and articles of clothing. 
We denote the under-determined SS followed by PR approach described in Section \ref{ssec:Sequential} with ``USS + PR''.  
We call the deep generative model based approach from Section \ref{sec:SSSPRwGen} ``Deep S$^3$PR''.

\paragraph{Measurement settings}
In each of our tests, the images have a resolution of $32\times32$ ($n=1024$) and we use $4\times$ oversampling ($m=4096$, $\mathbf{A}\in\mathbb{C}^{4096\times1024}$). For the case of Fourier measurements, we achieve the oversampling by first zero padding the $32\times 32$ images to $64\times 64$. We apply the algorithms to mixtures of two, three, and four images. We add white Gaussian noise to all our measurements, such that they have an SNR of $50$. Lower SNR results are included in the supplement. 

\begin{figure*}[h]
	\includegraphics[width=\textwidth]{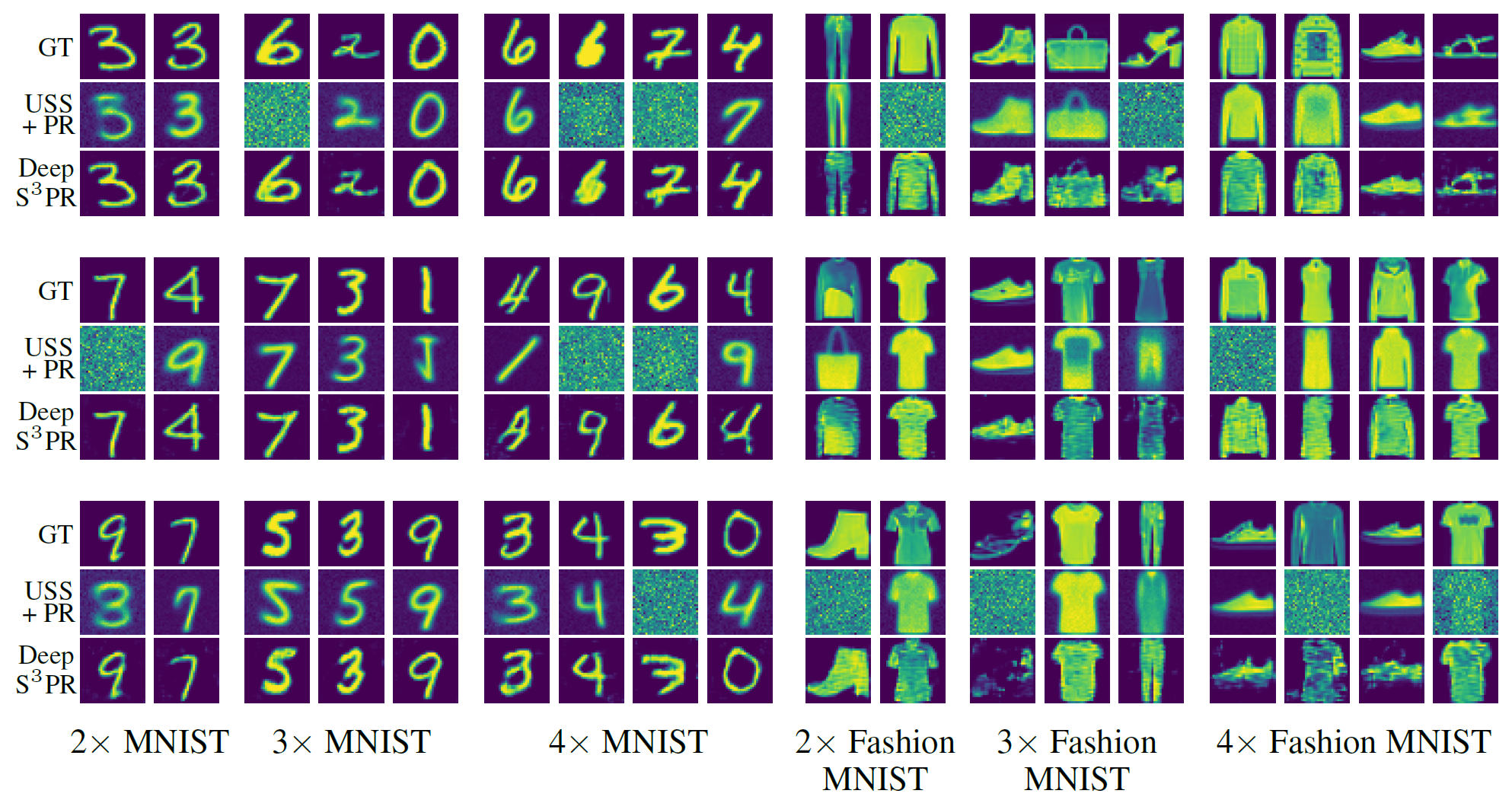}
	\caption{\textbf{Reconstructions from mixtures of two, three, or four images with a Gaussian measurement matrix}. With Gaussian measurements, the sequential algorithm (middle row of each set) can often recovery an image or two from the mixture, it almost always makes a few mistakes. In contrast, the proposed Deep S$^3$PR technique (bottom row of each set) largely succeeds in recovering up to four mixed images. Ground truth shown in the top row of each set.}
\label{fig:Gaussian}
\vspace{2mm}
\captionof{table}{Average NMSE across 10 sets of test images with Gaussian measurement matrices.}
\begin{tabular}{lcccccc}
	\toprule &   \shortstack{2$\times$ \\  MNIST}   &  \shortstack{3$\times$ \\  MNIST} & \shortstack{4$\times$ \\  MNIST} & \shortstack{2$\times$ Fashion\\  MNIST}   &  \shortstack{3$\times$ Fashion\\  MNIST} & \shortstack{4$\times$ Fashion\\  MNIST} \\
	\midrule
	USS + PR & $.54\pm.56$& $.52\pm.51$ & $.43\pm .32$&$.55\pm.37$ &$.40\pm.23$ &$.49\pm.14$\\
	Deep S$^3$PR & $.02\pm.02$&$.02\pm.02$ & $.04\pm.04$&$.08\pm.03$ &$.13\pm.09$ &$0.14\pm.07$\\
	\bottomrule
\end{tabular}
\label{tab:Gaussian}
\end{figure*}

\begin{figure*}[h]
	\includegraphics[width=\textwidth]{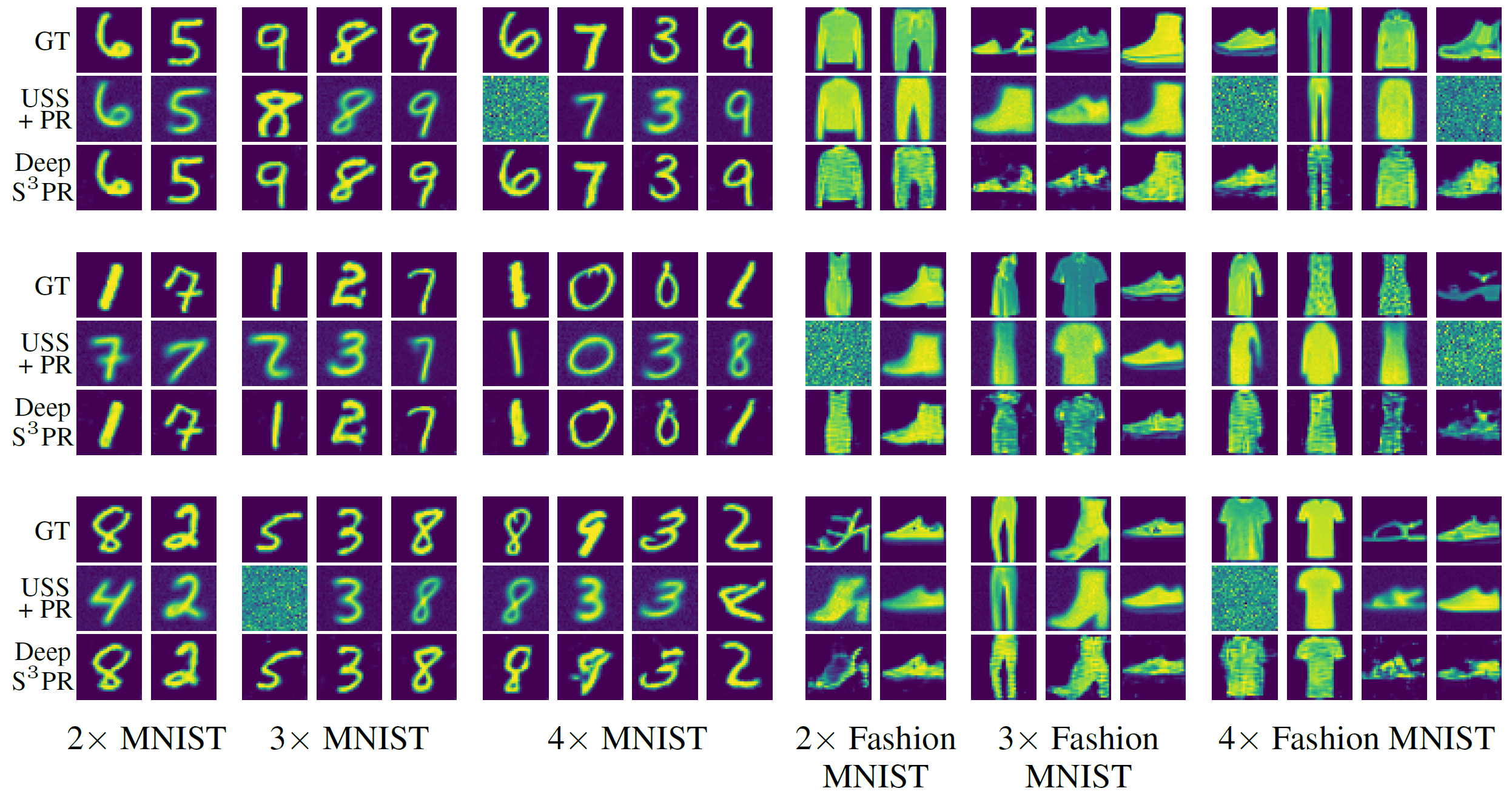}
	\caption{\textbf{Reconstructions from mixtures of two, three, or four images with a CDP measurement matrix}.  With CDP measurements the sequential algorithm (middle row of each set) often produces errors with just two images. In contrast the proposed Deep S$^3$PR method (bottom row of each set) again succeeds in recovering up to four mixed images. Ground truth shown in the top row of each set.}
\label{fig:CDP_results}
\vspace{2mm}
\captionof{table}{Average NMSE across 10 sets of test images with CDP measurement matrices.}
\begin{tabular}{lcccccc}
	\toprule &   \shortstack{2$\times$ \\  MNIST}   &  \shortstack{3$\times$ \\  MNIST} & \shortstack{4$\times$ \\  MNIST}  & \shortstack{2$\times$ Fashion\\  MNIST}   &  \shortstack{3$\times$ Fashion\\  MNIST} & \shortstack{4$\times$ Fashion\\  MNIST} \\
	\midrule
	USS + PR & $.40\pm.34$& $.33\pm.18$ & $.44\pm .15$&$.30\pm.31$ &$.60\pm.45$ &$.52\pm.29$\\
	Deep S$^3$PR & $.01\pm.01$&$.03\pm.03$&$.05\pm.03$&$.05\pm.03$&$.12\pm.09$&$.10\pm.04$\\
	\bottomrule
\end{tabular}
\label{tab:CDP}
\end{figure*}

\begin{figure*}[h]
\includegraphics[width=\textwidth]{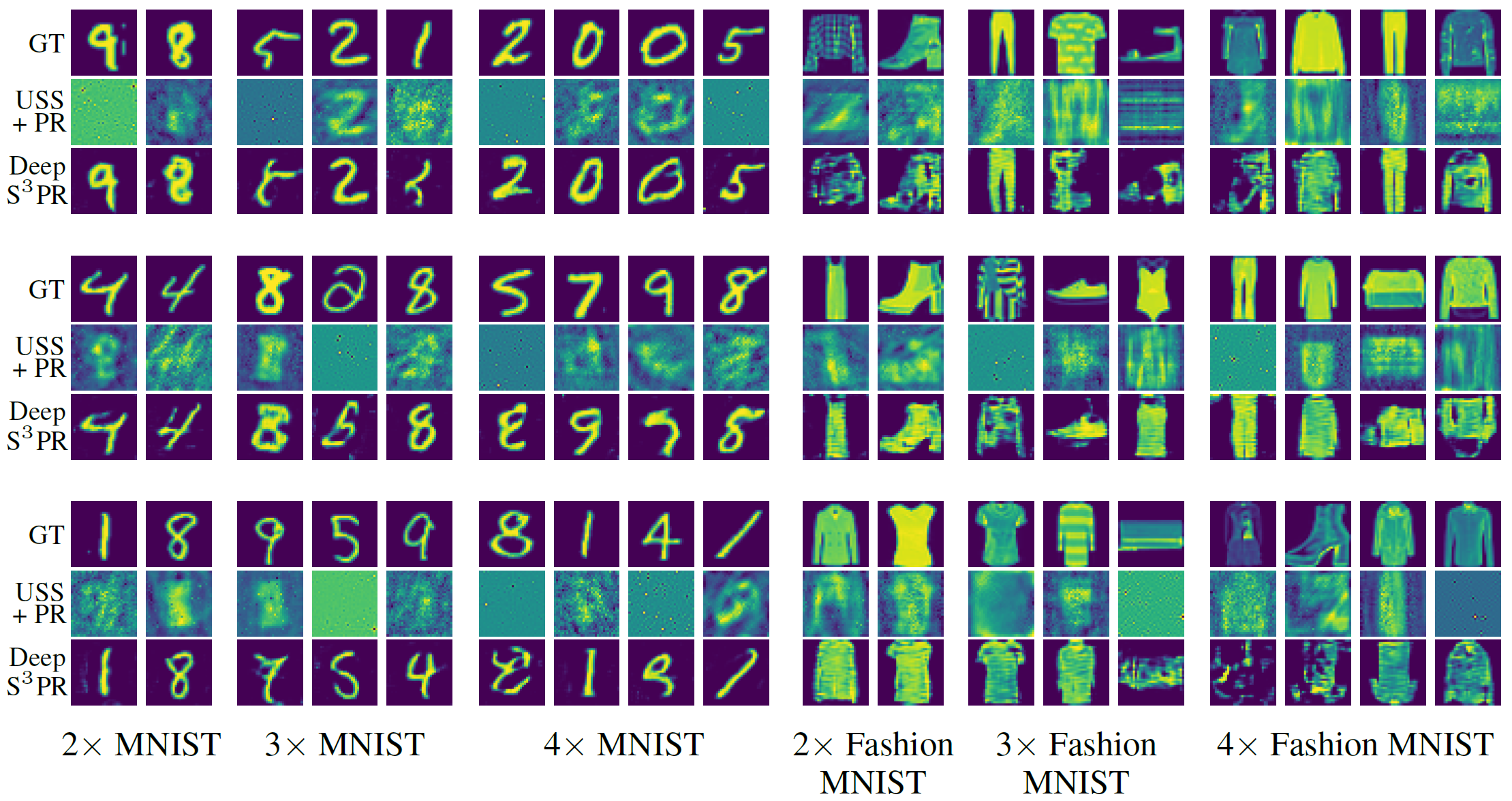}
\caption{\textbf{Reconstructions from mixtures of two, three, or four images with a Fourier measurement matrix}. Fourier measurements are particularly challenging. The sequential algorithm (middle row of each set) is unable to reconstruct almost any images images. The proposed Deep S$^3$PR method (bottom row of each set) recovers sets of two images but starts to make errors when tasked with recovering three or more images. With Fashion MNIST, noticeable reconstruction artifacts appear with just two images. Ground truth shown in the top row of each set.}
\label{fig:Fourier}
\vspace{2mm}
\captionof{table}{Average NMSE across 10 sets of test images with Fourier measurement matrices.}
\begin{tabular}{lcccccc}
	\toprule &  \shortstack{2$\times$ \\  MNIST}   &  \shortstack{3$\times$ \\  MNIST} & \shortstack{4$\times$ \\  MNIST} & \shortstack{2$\times$ Fashion\\  MNIST}   &  \shortstack{3$\times$ Fashion\\  MNIST} & \shortstack{4$\times$ Fashion\\  MNIST} \\
	\midrule
	USS + PR & $.68\pm.17$& $.83\pm.14$ & $.87\pm .12$&$.71\pm.21$ &$.79\pm.26$ &$.89\pm.12$\\
	Deep S$^3$PR & $.14\pm.09$&$.20\pm.06$&$.28\pm.06$&$.31\pm.21$&$.36\pm.16$ & $.42\pm.10$\\
	\bottomrule
\end{tabular}
\label{tab:Fourier}
\end{figure*}

We apply the algorithms to Gaussian, CDP, and Fourier measurements to generate Figures \ref{fig:Gaussian}, \ref{fig:CDP_results}, and \ref{fig:Fourier}, respectively. To generate each of the quantitative results presented in Tables \ref{tab:Gaussian}, \ref{tab:CDP}, and \ref{tab:Fourier} we apply the algorithms to 10 sets of images and compute the average normalized mean squared error (NMSE). To account for the labeling ambiguity, that is the solution $\hat{x}_1=A$, $\hat{x}_2=B$ is equivalent to the solution $\hat{x}_1=B$, $\hat{x}_2=A$, we report the loss associated with the ordering of the solutions that produces the minimum error. There is a sign ambiguity, that is $\hat{x}_1=A$ is an equivalent solution to $\hat{x}_1=-A$, that we similarly account for. 
Likewise, for Fourier measurements, we account for the flip ambiguities of the solutions by searching, over all flips left-right and up-down, for the one that minimizes the error. 

\paragraph{Algorithm settings}
Both the sequential USS + PR algorithm and Deep S$^3$PR method are implemented in Pytorch. Their respective optimization problems are solved with the ADAM optimizer with a learning rate of $0.02$. ADAM's momentum decay terms $b_1$ and $b_2$ are set to their default values of $0.9$ and $0.999$ respectively. 
For USS + PR, we minimize the under-determined SS loss~\eqref{eqn:SS_subproblem} by running the ADAM optimizer for $2\,000$ iterations. The PR loss~\eqref{eqn:PR_subproblem} is similarly minimized by running the ADAM optimizer for $2\,000$ iterations. 
For Deep S$^3$PR we minimize the loss~\eqref{eqn:Loss} by running the ADAM optimizer for $2\,000$ iterations. 
For both USS + PR and Deep S$^3$PR we perform 5 restarts.  For USS + PR, at each restart the estimates $\alpha$ and $x_1,~x_2,~\ldots,~x_L$ are initialized with random i.i.d.~Gaussian vectors with mean zero and unit variance. 
Similarly, for Deep S$^3$PR at each restart the latent vectors $z_1,~z_2,~\ldots,~z_L$ are initialized with random i.i.d.~Gaussian vectors with mean zero and unit variance. For both, we use the result with the smallest residual error, $\|y-\sum_l |\mathbf{A}\hat{x}_l|\|^2$, as our final solution.

\Subsection{Gaussian measurements}

\newcommand\mywidth{.6cm}

We first test the proposed S$^3$PR methods on complex-valued Gaussian measurement matrices. The elements of our measurement matrices are drawn from an i.i.d.~circular Gaussian distribution (the real and imaginary parts of each element are drawn from i.i.d.~distributions).
 
\paragraph{Results} 
Figure \ref{fig:Gaussian} demonstrates that Deep S$^3$PR is very effective with Gaussian measurement matrices. Even with mixtures of four images, Deep S$^3$PR produces near perfect reconstructions of MNIST digits and recognizable, though imperfect, reconstructions of Fashion MNIST articles of clothing as well. In contrast, the sequential solution to S$^3$PR produces significant errors with even just two images. The quantitative results, presented in Table \ref{tab:Gaussian}, mirror these findings.

\Subsection{Coded diffraction pattern measurements}

We next test our methods on simulated CDP measurements, which were first proposed in~\cite{candesCodedDiffraction}. The CDP measurement matrix can be written as \begin{align}
\mathbf{A}=\left[
\begin{array}{ll}
\mathbf{F}\mathbf{D}_1\\
\mathbf{F}\mathbf{D}_2\\
\mathbf{F}\mathbf{D_3}\\
\mathbf{F}\mathbf{D}_4
\end{array}
\right],
\end{align}
where $\mathbf{F}$ represents the two dimensional Fourier transform and $\mathbf{D}_1,~\mathbf{D}_2,~\mathbf{D}_3,$ and $\mathbf{D}_4$ are diagonal matrices whose diagonal entries are drawn uniformly from the unit circle in the complex plane. 

\paragraph{Results}
Figure~\ref{fig:CDP_results} and Table ~\ref{tab:CDP} demonstrate that Deep S$^3$PR is largely effective with CDP measurement matrices as well.   
Meanwhile, as it did with Gaussian measurements, the sequential solution to S$^3$PR makes errors starting with just two images.

\Subsection{Fourier measurements}\label{ssec:Fourier}

Finally, we test the proposed methods with Fourier measurements, which is arguably their most important use case. Fourier measurements form the basis of most coherence diffraction imaging systems~\cite{miao1999extending} as well as various correlation-based imaging systems~\cite{bertolotti2012non,katz2014non,metzlerOptica}. See the supplement for more information.

\paragraph{Preconditioning.}
When dealing with Fourier measurements, we minimize the loss \eqref{eqn:Loss} by minimizing
\begin{align}\label{eqn:Loss_w_corr}
\argmin_{z_1,z_2,\ldots,z_L} \|\mathbf{F}^{-1}y - \sum_{l=1}^L G(z_l)\star G(z_l)\|^2_2,
\end{align}
where $\star$ denotes autocorrelation. The equivalence of \eqref{eqn:Loss} and \eqref{eqn:Loss_w_corr} follows from Parseval's theorem and the relationship $\mathbf{F}(x\star x)=|\mathbf{F}x|^2$. We found this formulation of the problem offered faster convergence than directly minimizing~\eqref{eqn:Loss}.

\paragraph{Results}
Fourier measurements prove to be significantly more challenging than Gaussian or CDP measurements. Figure \ref{fig:Fourier} demonstrates that with three MNIST images, Deep S$^3$PR starts to make errors. While the technique can often reconstruct the general shape of Fashion MNIST images, there are artifacts in most reconstructions. Table \ref{tab:Fourier} shows these errors are reflected in the average NMSE as well. 
With Fourier measurements the sequential USS + PR algorithm fails with even two images.

%% file: sections/Discussion.tex
This work introduces and formalizes the simultaneous source separation and phase retrieval (S$^3$PR) problem. It then demonstrates how S$^3$PR can be solved by optimizing the latent variables of a deep generative model. We effectively apply the proposed Deep S$^3$PR technique to various mixtures of Gaussian, CDP, and Fourier measurements.

By leveraging the powerful image priors encoded in a pretrained deep generative model, Deep S$^3$PR is able to solve a problem that stymies classical, dictionary-based algorithms. 
Moreover, because the generative model is problem agnostic, the proposed method generalizes across forward models, mixtures sizes, noise levels and more. This stands in stark contrast to a discriminative neural network approach, which would need to be retrained each time one of these parameters changes.

Finally, Deep S$^3$PR represents a major step forward in addressing a long-standing challenge in computational optics and we look forward to experimentally validating its benefits with physical experiments; such as imaging extended objects through scattering media~\cite{bertolotti2012non,katz2014non} or around corners~\cite{viswanath2018indirect,metzlerOptica}. 
Deep S$^3$PR stands to enable major strides in these and other applications.

%% file: sections/Supp_Sec_simplefigs.tex
\begin{abstract}
This supplement provides a detailed description of how simultaneous source separation and phase retrieval (S$^3$PR) appears and limits performance in a real-world computational optics application; seeing through scattering media. 
It also provides additional experimental results tested at a lower signal-to-noise ratio (SNR).
\end{abstract}

\section{S$^3$PR in computational optics}
\label{sec:S3PR_in_optics}
Phase retrieval (PR) is a fundamental part of many computational optics/imaging systems. For instance, it is used in microscopy to exceed the diffraction limit~\cite{maiden2011superresolution}, X-ray crystallography to perform imaging without a lens~\cite{miao1999extending}, and astronomical imaging to see through atmospheric aberrations~\cite{elbaum1974laser}.

Implicit in many of these systems is an assumption that there is a {\em single} coherent illumination source. (A source is said to be coherent if its constituent fields maintain a constant phase offset, and thus produce construct and destructive interference.) When this assumption is broken, and there are instead multiple independent coherent sources, the reconstruction problem becomes S$^3$PR rather than PR. Accordingly, our inability to effectively solve S$^3$PR places practical limitations on the performance of these systems. 

We next describe the physics and mathematics behind a real-world imaging systems where S$^3$PR limits performance. 
Similar limits play out in many other imaging applications.

\subsection{Speckle correlation imaging through scattering media}

Speckle correlation imaging through scattering media was first introduced in 2012~\cite{bertolotti2012non} and has since been extended in~\cite{katz2014non} and many other works. It is based off of ideas first developed for astronomical imaging nearly 50 years ago~\cite{labeyrie1970attainment}. 

\begin{figure}
	\centering
	\begin{subfigure}[t]{.8\textwidth}
	\includegraphics[width=\textwidth]{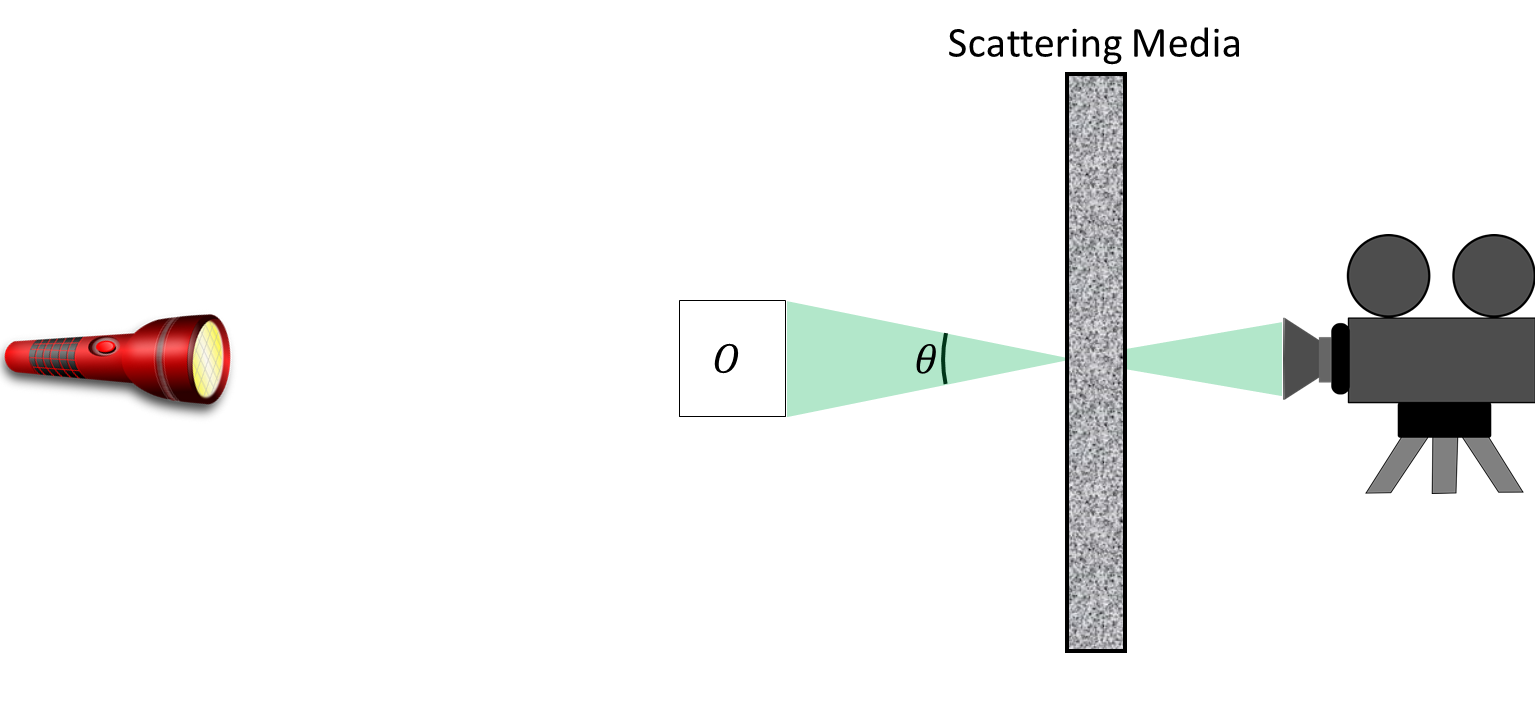}
	\caption{One object spanning a limited field-of-view}
	\label{fig:OpticalSetup_oneO}
	\end{subfigure}
	\begin{subfigure}[t]{.8\textwidth}
	\includegraphics[width=\textwidth]{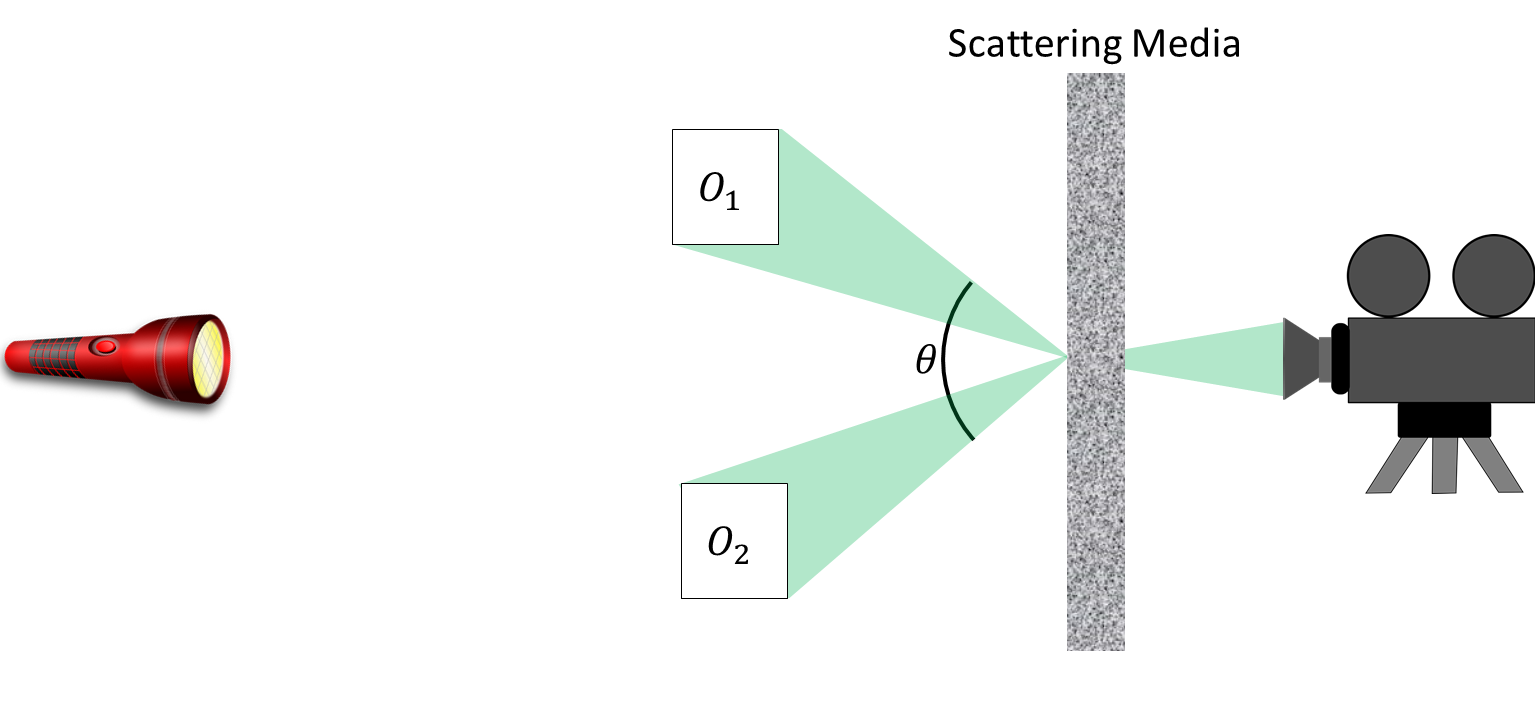}
	\caption{Two objects spanning an extended field-of-view}
	\label{fig:OpticalSetup_twoO}
	\end{subfigure}
	\caption{{\bf Speckle correlation imaging optical configuration.} Until now, correlation-based imaging systems have only been able to image a single object that subtends an angle, $\theta$, that is smaller than the range of the memory-effect. A solution to S$^3$PR opens up the possibility of imaging multiple, far-apart objects, which together subtend an angle greater than the memory-effect range.}
	\vspace{-15 pt}
\end{figure}

This family of techniques allows one to non-invasively image simple objects (\cite{katz2014non} reconstructed phantoms consisting of letters, numbers, and smiley faces) through thin scattering media, like a layer of soft tissue. The technique can reconstruct millimeter-scale features and requires only a temporally coherent, spatially incoherent light source and a camera~\cite{katz2014non}. The physical setup associated with speckle correlation imaging is illustrated in Figure~\ref{fig:OpticalSetup_oneO}.

Speckle correlation imaging through scattering media is based on a physical phenomenon known as the angular memory-effect~\cite{freund1988memory}. It states that if an object $O$ illuminates a scattering media with a temporally coherent light, each point of the scattering media will produce a {\em spatially invariant} interference pattern $S$ (known as ``speckle''), so long as, from the perspective of the scattering media, {\em the angle subtended by the object is small.}

Because of the memory-effect--induced spatial invariance of the speckle, the measurement model associated with Figure~\ref{fig:OpticalSetup_oneO} is convolutional and is described by
\begin{align}
I = S * O,
\end{align}
where $S$ is the unknown speckle pattern, $O$ is the object of interest, and $*$ denotes convolution.

Because the autocorrelation function of a speckle pattern is Dirac-like~\cite{goodman2007speckle}, ignoring DC terms we have that
\begin{align}
I\star I &= (S * O) \star (S * O)\nonumber,\\
&= (S\star S) * (O\star O)\nonumber,\\
& \approx O\star O,
\end{align}
where $\star$ denotes correlation.

With this estimate of the autocorrelation of the hidden object $O$ in hand, we can reconstruct the object by using phase retrieval algorithms and the relationship
\begin{align}
\mathcal{F}(O\star O) = |\mathcal{F}(O)|^2,
\end{align}
where $\mathcal{F}(\cdot)$ denotes the Fourier transform operator.

\subsection{Speckle correlation imaging over an extended field-of-view}

Now consider two small, spatially separated objects, $O_1$ and $O_2$, as illustrated in Figure~\ref{fig:OpticalSetup_twoO}. Each object itself subtends a small angle, and thus experiences spatially invariant speckle. However, they are far apart from one another and so each experiences a different speckle pattern. Thus, the measurement formation model is described by
\begin{align}
I = S_1 * O_1 + S_2 * O_2,
\end{align}
where $S_1$ and $S_2$ denote two independent speckle realizations.

Because the two speckle realizations are uncorrelated, again ignoring DC terms, $S_1\star S_2\approx0$. Thus, the autocorrelation of $I$ becomes
\begin{align}
I\star I  =& ( S_1 * O_1 + S_2 * O_2) \star ( S_1 * O_1 + S_2 * O_2),\nonumber\\
 =& (S_1\star S_1) * (O_1\star O_1) + (S_2\star S_2) * (O_2\star O_2) \nonumber\\
 &+ ( S_1 * O_1)\star( S_2 * O_2) + ( S_2 * O_2)\star( S_1 * O_1),\nonumber\\
\approx& (S_1\star S_1) * (O_1\star O_1) + (S_2\star S_2) * (O_2\star O_2),\nonumber\\
\approx& O_1\star O_1+ O_2\star O_2.
\end{align}

Taking the Fourier transform of the result, we arrive at
\begin{align}
\mathcal{F}(I\star I) & = \mathcal{F}(O_1\star O_1+ O_2\star O_2), \nonumber\\
& = \mathcal{F}(O_1\star O_1)+ \mathcal{F}(O_2\star O_2),\nonumber\\
& = |\mathcal{F}(O_1)|^2+ |\mathcal{F}(O_2)|^2.
\end{align}

In this way, speckle correlation imaging over an extended field of view naturally leads to an S$^3$PR reconstruction problem. 
Until now, correlation-based imaging through scattering media systems have avoided or ignored S$^3$PR by either (1) restricting themselves to imaging only a single small object\cite{katz2014non}, (2) using invasive illumination sources that illuminate only one portion of a hidden object at a time~\cite{gardner2019ptychographic}, or (3) capturing thousands of training images and throwing a discriminative neural network at the problem~\cite{li2018deep}. 
Our work makes strides towards removing these limitations.

\section{Additional Results}
\label{sec:More_results}

\begin{figure*}
	\includegraphics[width=\textwidth]{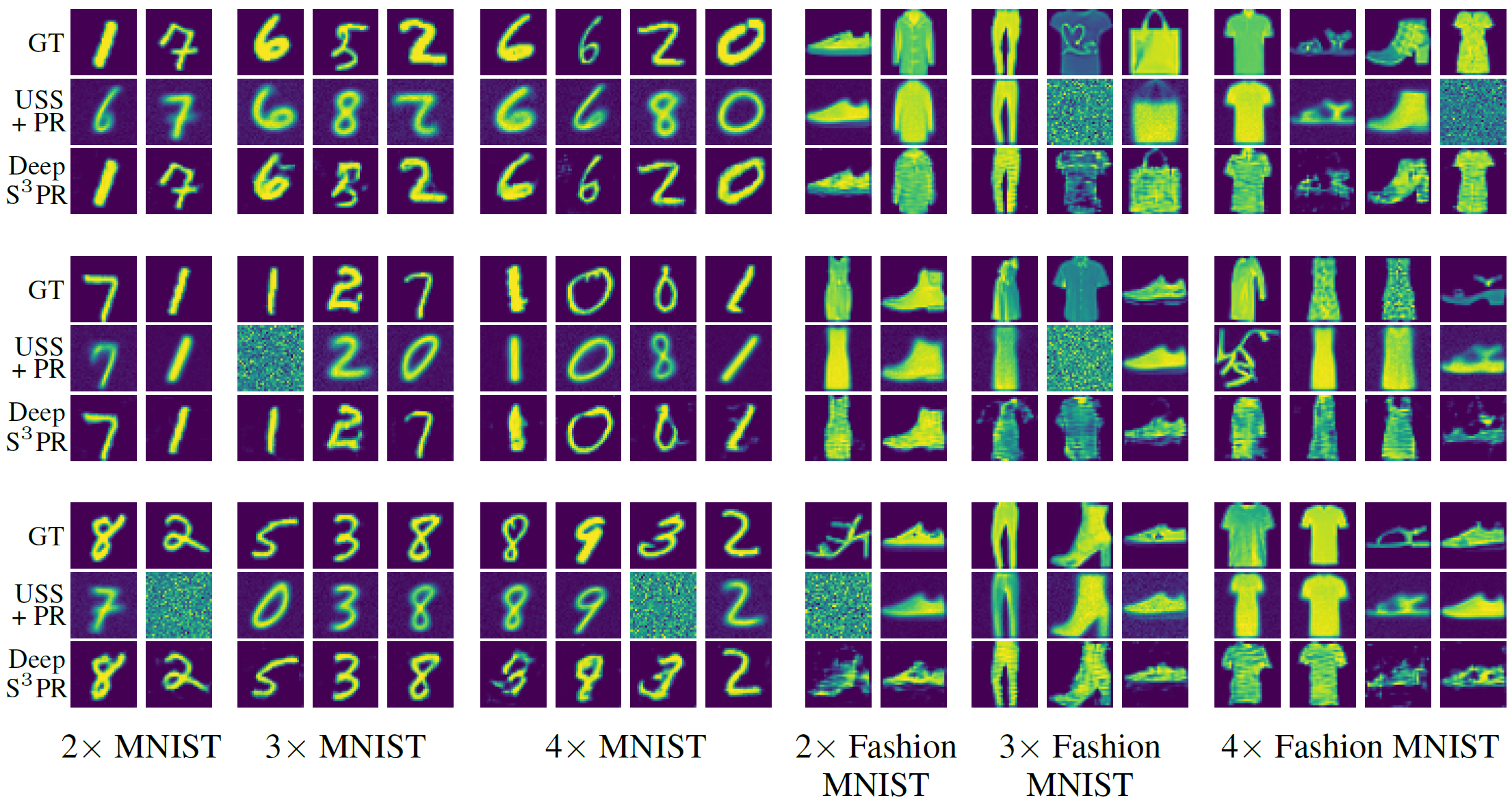}
	\caption{\textbf{Gaussian Measurement Matrix Results}. Top: Ground truth. Middle: USS+PR reconstructions. Bottom: Deep S$^3$PR reconstructions.}
\label{fig:Gaussian2}
\vspace{2mm}
\captionof{table}{Average NMSE across 10 sets of test images with Gaussian measurement matrices.}
\begin{tabular}{lcccccc}
	\toprule &   \shortstack{2$\times$ \\  MNIST}   &  \shortstack{3$\times$ \\  MNIST} & \shortstack{4$\times$ \\  MNIST}  & \shortstack{2$\times$ Fashion\\  MNIST}   &  \shortstack{3$\times$ Fashion\\  MNIST} & \shortstack{4$\times$ Fashion\\  MNIST} \\
	\midrule
	USS + PR & $.37\pm.34$& $.37\pm.32$ & $.45\pm .35$&$.46\pm.44$ &$.46\pm.28$ &$.34\pm.19$\\
	Deep S$^3$PR & $.01\pm.01$&$.04\pm.04$ & $.09\pm.06$&$.05\pm.03$ &$.14\pm.08$ &$0.12\pm.04$\\
	\bottomrule
\end{tabular}
\label{tab:Gaussian2}
\end{figure*}

\begin{figure*}
	\includegraphics[width=\textwidth]{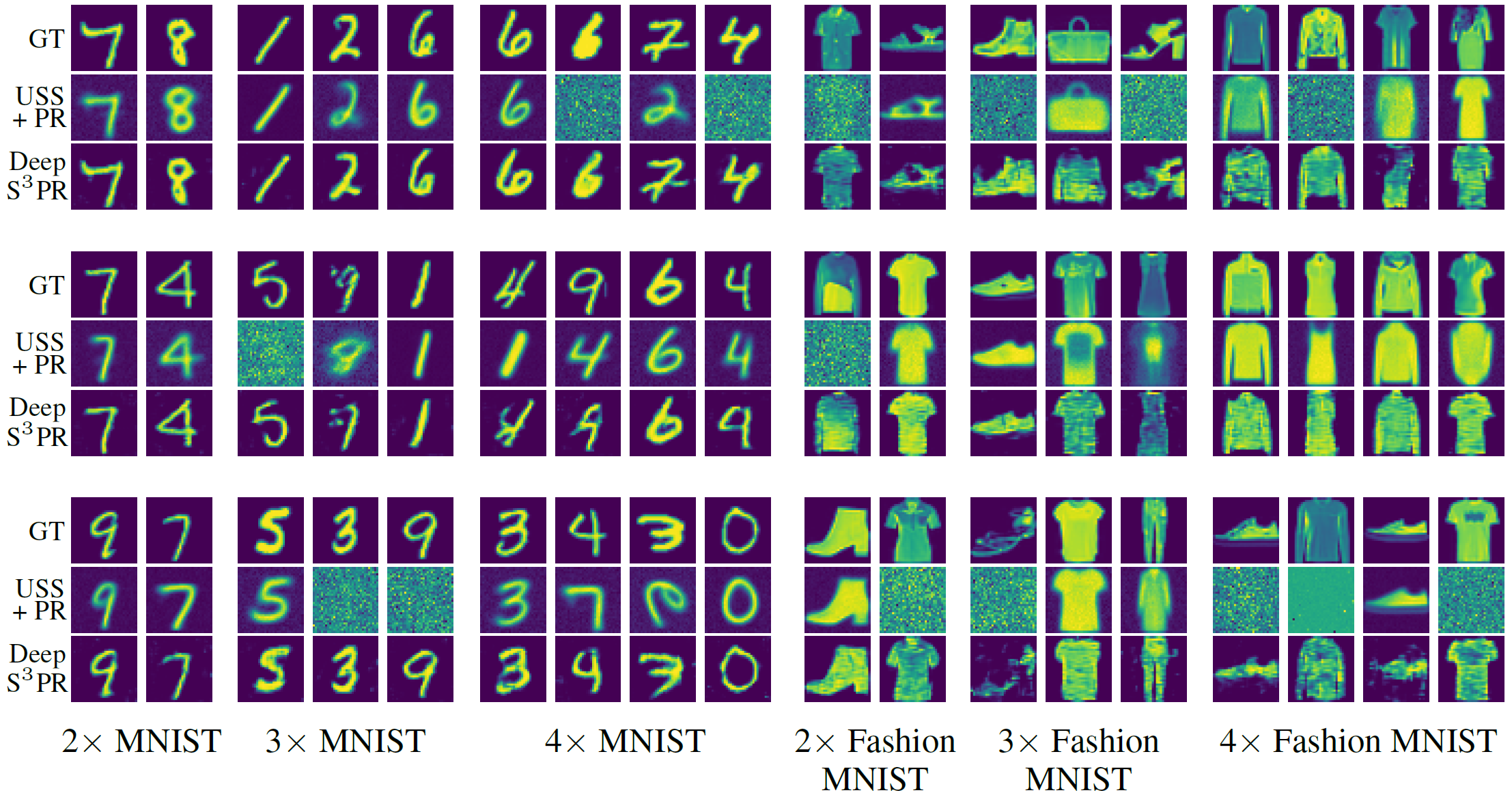}
		\caption{\textbf{Coded Diffraction Pattern Measurement Matrix Results}.  Top: Ground truth. Middle: USS+PR reconstructions. Bottom: Deep S$^3$PR reconstructions.}
	\label{fig:CDP_results2}
	\vspace{2mm}
	\captionof{table}{Average NMSE across 10 sets of test images with CDP measurement matrices.}
	\begin{tabular}{lcccccc}
		\toprule &   \shortstack{2$\times$ \\  MNIST}   &  \shortstack{3$\times$ \\  MNIST} & \shortstack{4$\times$ \\  MNIST}  & \shortstack{2$\times$ Fashion\\  MNIST}   &  \shortstack{3$\times$ Fashion\\  MNIST} & \shortstack{4$\times$ Fashion\\  MNIST} \\
		\midrule
		USS + PR & $.15\pm.05$& $.45\pm.41$ & $.46\pm .30$&$.71\pm.53$ &$.40\pm.40$ &$.50\pm.24$\\
		Deep S$^3$PR & $.01\pm.01$&$.02\pm.01$&$0.05\pm.03$&$.09\pm.04$&$.14\pm.08$&$.15\pm.06$\\
		\bottomrule
	\end{tabular}
	\label{tab:CDP2}
\end{figure*}

\begin{figure*}
	\includegraphics[width=\textwidth]{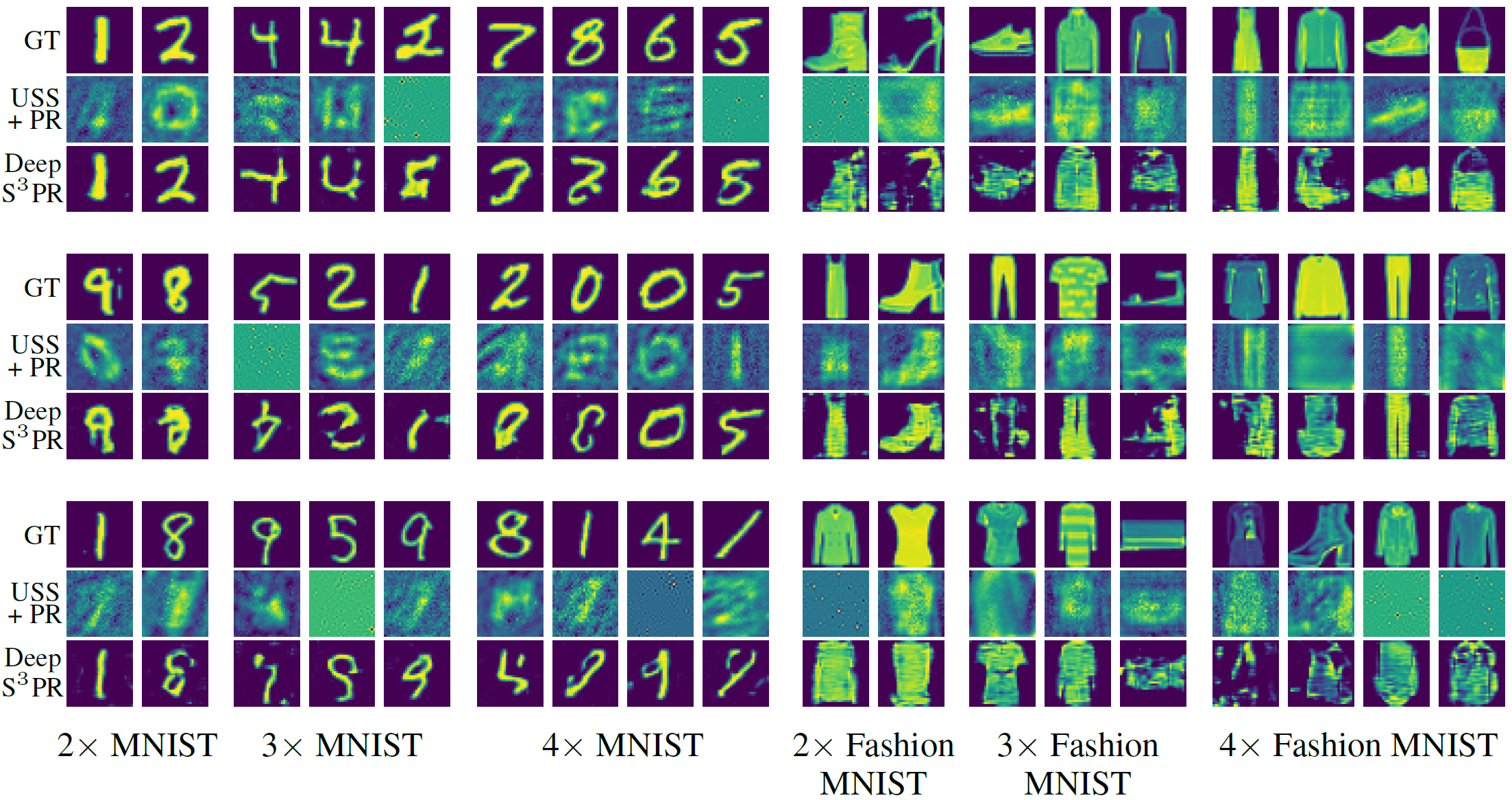}
	\caption{\textbf{Fourier Measurement Matrix Results}. Top: Ground truth. Middle: USS+PR reconstructions. Bottom: Deep S$^3$PR reconstructions.}
\label{fig:Fourier2}
\vspace{2mm}
\captionof{table}{Average NMSE across 10 sets of test images with Fourier measurement matrices.}
\begin{tabular}{lcccccc}
	\toprule &  \shortstack{2$\times$ \\  MNIST}   &  \shortstack{3$\times$ \\  MNIST} & \shortstack{4$\times$ \\  MNIST} & \shortstack{2$\times$ Fashion\\  MNIST}   &  \shortstack{3$\times$ Fashion\\  MNIST} & \shortstack{4$\times$ Fashion\\  MNIST} \\
	\midrule
	USS + PR & $.60\pm.15$& $.82\pm.11$ & $.78\pm .12$&$.68\pm.24$ &$.82\pm.18$ &$.90\pm.15$\\
	Deep S$^3$PR & $.18\pm.08$&$.26\pm.04$&$.29\pm.03$&$.28\pm.14$&$.40\pm.16$ & $.41\pm.18$\\
	\bottomrule
\end{tabular}
\label{tab:Fourier2}
\end{figure*}

Figures~\ref{fig:Gaussian2},~\ref{fig:CDP_results2}, and~\ref{fig:Fourier2} and Tables~\ref{tab:Gaussian2},~\ref{tab:CDP2}, and~\ref{tab:Fourier2} provide additional experimental results, captured at an SNR of 15 rather than 50. The resulting Gaussian and coded diffraction pattern (CDP) reconstructions are similar to the higher SNR reconstructions from the main text, while the low SNR Fourier results exhibit a few more artifacts.

